\documentclass[letterpaper]{article} 
\usepackage{aaai24}  
\usepackage{times}  
\usepackage{helvet}  
\usepackage{courier}  
\usepackage[hyphens]{url}  
\usepackage{graphicx} 
\usepackage{natbib}  
\usepackage{caption} 
\usepackage{algorithm}
\usepackage{algorithmic}

\usepackage{newfloat}
\usepackage{listings}
\DeclareCaptionStyle{ruled}{labelfont=normalfont,labelsep=colon,strut=off} 
\floatstyle{ruled}
\newfloat{listing}{tb}{lst}{}
\floatname{listing}{Listing}
\title{Learning Visual Abstract Reasoning through Dual-Stream Networks}
\author{
    Kai Zhao\textsuperscript{\rm 1},
    Chang Xu\textsuperscript{\rm 1},
    Bailu Si\textsuperscript{\rm 1, 2}\thanks{Corresponding author.}
}
\affiliations{
    \textsuperscript{\rm 1}School of Systems Science, Beijing Normal University\\
    \textsuperscript{\rm 2}Chinese Institute for Brain Research, Beijing
    
    {\rm zhaokai\_id@foxmail.com, changxu@mail.bnu.edu.cn, bailusi@bnu.edu.cn}
}

\usepackage{bibentry}

\begin{document}

\maketitle

\begin{abstract}
Visual abstract reasoning tasks present challenges for deep neural networks, exposing limitations in their capabilities. In this work, we present a neural network model that addresses the challenges posed by Raven's Progressive Matrices (RPM). Inspired by the two-stream hypothesis of visual processing, we introduce the Dual-stream Reasoning Network (DRNet), which utilizes two parallel branches to capture image features. On top of the two streams, a reasoning module first learns to merge the high-level features of the same image. Then, it employs a rule extractor to handle combinations involving the eight context images and each candidate image, extracting discrete abstract rules and utilizing an multilayer perceptron (MLP) to make predictions.
Empirical results demonstrate that the proposed DRNet achieves state-of-the-art average performance  across multiple RPM benchmarks. Furthermore, DRNet demonstrates robust generalization capabilities, even extending to various out-of-distribution scenarios. The dual streams within DRNet serve distinct functions by addressing local or spatial information. They are then integrated into the reasoning module, leveraging abstract rules to facilitate the execution of visual reasoning tasks. These findings indicate that the dual-stream architecture could play a crucial role in visual abstract reasoning.
\end{abstract}

\section{Introduction}  
One goal of artificial intelligence (AI) is to equip machines with universal reasoning capabilities.  Presently, deep learning has emerged as the dominant paradigm in AI, enabling the modeling of data to execute intricate tasks such as image classification \cite{he2016deep, dosovitskiy2020image}, object recognition  \cite{girshick2015region, ronneberger2015u}, and natural language processing \cite{vaswani2017attention}.  In the field of cognitive science, analogical reasoning has consistently been regarded as the foundation of general intelligence that sets humans apart from animals and is considered the essence of cognition.  It is often shaped by the interplay between higher cognitive abilities and the quality of incoming representations (Norman, 1975).  However, current deep learning systems still struggle to excel in tasks that demand analogical and relational reasoning.

\begin{figure}[t]
    \centering
    \includegraphics[scale=0.43]{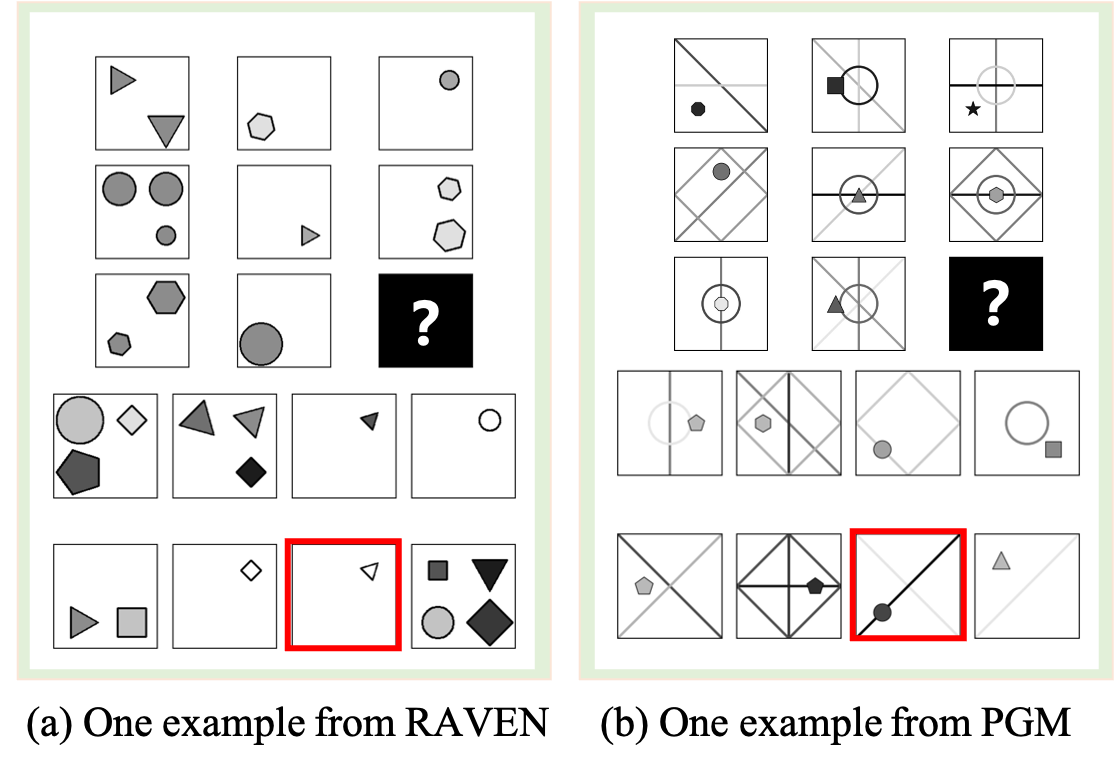}
    \caption{Examples from RAVEN and Procedurally Generated Matrices (PGM) are shown in (a) and (b) respectively. Both types of problems involve presenting participants with eight context images. They are required to select the correct answer (highlighted in red) from the candidate set of eight images to fill in the blank (denoted by \textbf{?}), in order to satisfy specific rules in the row or column direction of the 3$\times$3 matrix.} 
    \label{fig:example}
\end{figure}

As a significant assessment tool in the realm of analogy reasoning, the RAVEN test boasts an 80-year history \cite{Court1982manual, Prabhakaran1997neural, perfetti2009differential}. Researchers from various disciplines, including psychology, cognitive science, and artificial intelligence, have extensively explored this area.
In recent years, preceding works \cite{zhang2019raven, hu2020stratified, benny2021scale} have generated program-controlled RPM datasets, as depicted in Figure \ref{fig:example}. These efforts have greatly facilitated deep learning research in this domain and have assessed the analogy reasoning capability of deep learning systems \cite{malkinski2022deep}.
Serving as a widely accepted benchmark for intelligence evaluation, RPM problem requires participants to identify one or more rules within a 3$\times$3 matrix, and then make a correct choice. This process of abstract reasoning mirrors the attributes of advanced human intelligence \cite{snow1984toward, snow1984topography, jaeggi2008improving}.

Inspired by the two-stream hypothesis \cite{goodale1992separate, grezes2002does, maguire1999human} in neuroscience, we propose a Dual-stream Reasoning Network (DRNet) to address the RPM problems.  
DRNet simulates object recognition through the ventral stream and spatial attention through the dorsal stream in the context of dual-stream vision. It extracts high-level visual features from these two streams. After performing fusion on the extracted visual features, DRNet feeds these features into the rule extractor to infer relationships between images, resulting in abstract rule representations. DRNet utilizes these rule representations to predict the correct answers.  Codes are available at https://github.com/VecchioID/DRNet. 
\begin{figure*}[h]
    \centering
    \includegraphics[scale=0.16]{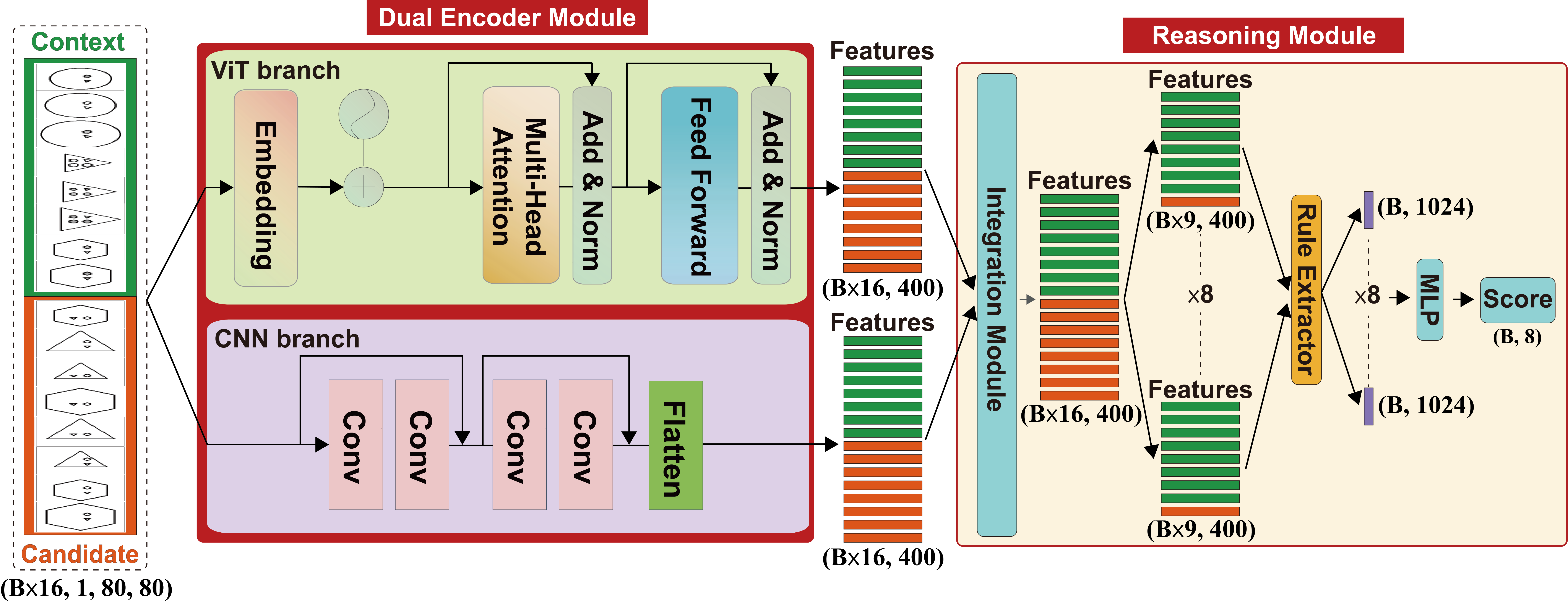}
    \caption{An overview of our DRNet. DRNet consists of a dual encoder module and a reasoning module, where (B$\times$16, 1, 80, 80) represents (batchsize$\times$16, channels, image size, image size). The dual encoder module is used to extract input image features in parallel, after which the features are fed into the reasoning module. The reasoning module first learns to merge the high-level embeddings of the same image. Then, it employs a rule extractor to handle combinations involving the eight context images and each candidate image, extracting abstract rules and utilizing an MLP to make predictions.
    } 
    \label{fig:framework}
\end{figure*}

We conduct comprehensive empirical studies on several RPM benchmarks. To summarize, our contributions include:
\begin{itemize}
    \item Unlike previous single-stream frameworks, DRNet combines the advantages of local and spatial representations, allowing it to exhibit distinct interpretations of input images. This collective enhancement improves the model's reasoning performance.
    
    \item DRNet achieves remarkable generalization performance and outperforms other models on multiple datasets, showcasing the effectiveness of this framework for non-verbal visual abstract reasoning problems.
    
    \item Visualization results of the rule representations indicate that the learned representations can be clustered based on rule categories, thereby facilitating visual abstract tasks.
\end{itemize}

\section{Related Work}

\subsection{Raven's Progressive Matrices}

Most previous work on RPMs has focused on single-stream network frameworks, such as ConvNets with inductive bias and ViTs focused on self-attention.

Early attempts at deep learning for RPM used a convolutional neural network (CNN) deep learning model by Hoshen and Werman \cite{hoshen2017iq}. 
Modern architectures such as WReN \cite{santoro2018measuring}, CoPINet \cite{zhang2019learning}, Rel-Base \cite{spratley2020closer}, SRAN \cite{hu2020stratified}, MRNet \cite{benny2021scale}, PredRnet \cite{pmlr-v202-yang23r}, etc. use variants of convolutional neural networks \cite{lecun1998gradient, he2016deep} for feature extraction. This suggests that the inductive bias can potentially generalize well in different configurations \cite{santoro2017simple, jahrens2020solving, zhuo2020effective, zhang2022learning, mondallearning, malkinski2022multi}.  Among all previous studies, SCL \cite{wu2020scattering} uses the compositional representation of object attributes and their relations for reasoning.
Some symbolically inspired models to incorporate logical rule or object vectors into the unidirectional flow framework, e.g. PrAE \cite{zhang2021abstract}, ALANS learner \cite{zhang2021learning} and NVSA \cite{hersche2023neuro}.

Another line of work focuses on attention mechanisms \cite{hahne2019attention, rahaman2021dynamic, mondallearning, PritishSAViR-T, xiaojianrelvit2022}. A recent study shows that visual transformers retain more spatial information than CNNs \cite{raghu2021vision}. Previous studies such as dynamic inference with neural interpreters~\cite{rahaman2021dynamic} and STSN \cite{mondallearning} explored the modular network architecture and image representations for abstract visual reasoning.

There are also some non-single stream studies, such as graph neural networks and reinforcement learning.
MXGNet \cite{wang2020abstract} proposes a multilayer graph neural network for multi-panel diagrammatic reasoning tasks, while LEN \cite{zheng2019abstract} demonstrates a reinforcement learning teacher model to guide the training process.

\subsection{Two Stream Networks}

Inspired by the two stream hypothesis, two stream networks are widely explored in the video action recognition field \cite{simonyan2014two, carreira2017quo, feichtenhofer2016convolutional, zolfaghari2017chained}.  I3D \cite{carreira2017quo} builds a two stream 3D-CNN  architecture and takes RGB video and optical flow as inputs. SlowFast \cite{feichtenhofer2019slowfast} is a model that encodes videos with different frame rates.  DS-Net \cite{mao2021dual} uses a dual-stream network to explore the representation capacity of local and global pattern features for image classification. \citeauthor{chen2022two} employ a dual visual encoder  containing two separate streams to model both the raw videos and the key-point sequences for sign language understanding \cite{chen2022two}. 
We introduce dual-stream networks to abstract visual reasoning tasks.  Although our work uses dual-stream networks like previous studies,  we have a different implementation.  First, dual-stream network was implemented as a dual encoder module to extract high level image features across different images. Second, we introduce a reasoning module in DRNet, which allows it to extract rules from different RPM problems.

How to model the interactions between different streams is non-trivial. I3D \cite{carreira2017quo} uses a late fusion strategy by
simply averaging the predictions of two streams. Another way is to fuse the intermediate features of
each stream in the early stage by lateral connections \cite{feichtenhofer2019slowfast}, concatenation \cite{zhou2021spatial}, or addition \cite{cui2019deep}.
In this work, our approach directly models local and spatial features via a dual encoder, and then a learnable module is used to fuse the intermediate features of each stream.

\section{Methods}
The structure of DRNet is shown in Figure \ref{fig:framework}. It consists of two components: (1) a dual encoder module to transform each image into two high-level features, and (2) a reasoning module to score each context candidate group's features. The highest score is selected as the final predicted answer.

\subsection{Dual Encoder Module} 
This module consists of two parallel streams, where a CNN is used to recognize objects to acquire local features, while ViT is used to play a role in attending to the spatial location of objects.

\textbf{CNN branch.} Our CNN stream has two ResBlocks, each containing a residual branch and a shortcut connection. Each residual branch has two convolutional layers with kernel sizes of 7. Each convolutional layer down-samples the input feature with a stride of 2, which expands the receptive fields of the neurons and allows for the extraction of higher-level information. The shortcut connection applies the $MaxPool2d$ operation twice to match the output size of the residual branch with a stride of 2. In total, our first ResBlock can be formulated as:
\begin{equation}
    x^{l} = ReLU(BN(Conv_{7\times7}(x^{l-1}))), l\in{1,2}
\end{equation}
\begin{equation}
    x^{l} = x^{l} + Maxpool2d(x^{0}), l=2
\end{equation}
where $x^{0}$ represent input features, $l$ represents the layer index of convolutional layers. 
$x^{cnn\_out}$ can be obtained by treating $x^{2}$ in the same way as above through the second ResBlock.
We set the filters as [64, 64, 64, 16] from the first to the last convolutional layer.

\textbf{ViT branch.}
ViT branch processes each image parallelly. Our ViT has the same network framework as in \cite{vaswani2017attention}, except that we employ 1D learnable positional encodings to add them to patch embeddings for retaining positional information. 
Our ViT has 8 attentional heads with depth of 12. 
We first split each image into 16 patches, with each patch size of 20$\times$20. One convolutional layer with a kernel of 20 and a stride 20, is applied to transform into a patch embedding with size of 400. After transformer encoder, we finally obtained an averaged feature vector. The ViT branch can be formulated as:
\begin{equation}
    x^{vit\_out} = ViT(x^{0}) 
\end{equation}
where $x^{vit\_out}$  represents the output features of ViT branch.

To clarify the process, we describe the data flow illustrated in Figure \ref{fig:framework}. For each RPM problem, we have 8 context images $I^{c}_{i}$, where $i\in\left\{1,2,...,8\right\}$, and 8 candidate images $I^{a}_{i}$, where $i\in\left\{1,2,...,8\right\}$, which are combined to create an input denoted as $I = [I^{c}_{1}, I^{c}_{2}, ...,I^{c}_{8}, I^{a}_{1}, I^{a}_{2}, ..., I^{a}_{8}]$, with $I\in\mathcal{R}^{(16\times 1\times 80 \times 80)}$. This input $I$ is then simultaneously fed into both the CNN branch and the ViT branch with a batch approach. The result is image features: $x^{cnn\_out}\in\mathcal{R}^{(B\times16, 1, 20, 20)}$ from the CNN branch and $x^{vit\_out}\in\mathcal{R}^{(B\times16, 400)}$ from the ViT branch. To ensure compatibility, we reshape $x^{cnn\_out}$ to $\mathcal{R}^{(B\times16, 400)}$, aligning its shape with that of $x^{vit\_out}$. Here, $B$ represents the batch size. Finally, the outputs $x^{vit\_out}$ and $x^{cnn\_out}$ are passed to the reasoning module.

\subsection{Reasoning Module} 
This reasoning module consists of an integration module and a rule extractor to fuse high-level features and extract abstract rule representations of RPM problems.

\textbf{Integration Module.} Just as the two streams ultimately project to the hippocampus \cite{huang2021extensive}, DRNet designs an integration module to model the interactions between different streams.
To promote order-invariance between the two vectors, a permutation-invariant operator is recommended. One can use the sum operator ($\mathbf{SUM}$). This approach has been employed by \cite{niebur1995control, benny2021scale}.
\begin{equation}
    \mathbf{SUM}(\cdot) := x^{cnn\_out} + x^{vit\_out}
\end{equation}

To reduce the variance of $\mathbf{SUM}$, a mean ($\mathbf{MEA}$) operator is defined as follows:
\begin{equation}
    \mathbf{MEA}(\cdot) := (x^{cnn\_out} + x^{vit\_out})/2
\end{equation}

As a variation of the above operators, we propose an adaptive attention operator $\mathbf{AUT}$ to automatically combine two streams, 
\begin{equation}
    \mathbf{AUT}(\cdot) := w_{1}x^{cnn\_out} + w_{2}x^{vit\_out}
\end{equation}
Where $w_{1}$ and $w_{2}$ are learnable tensors.
We provide three methods for determining changes in these two parameters: $\mathbf{AUT-L1}$ normalization, $\mathbf{AUT-L2}$ normalization, and the unrestricted way ($\mathbf{AUT}$).

\begin{figure}[b]
    \centering
    \includegraphics[scale=0.2]{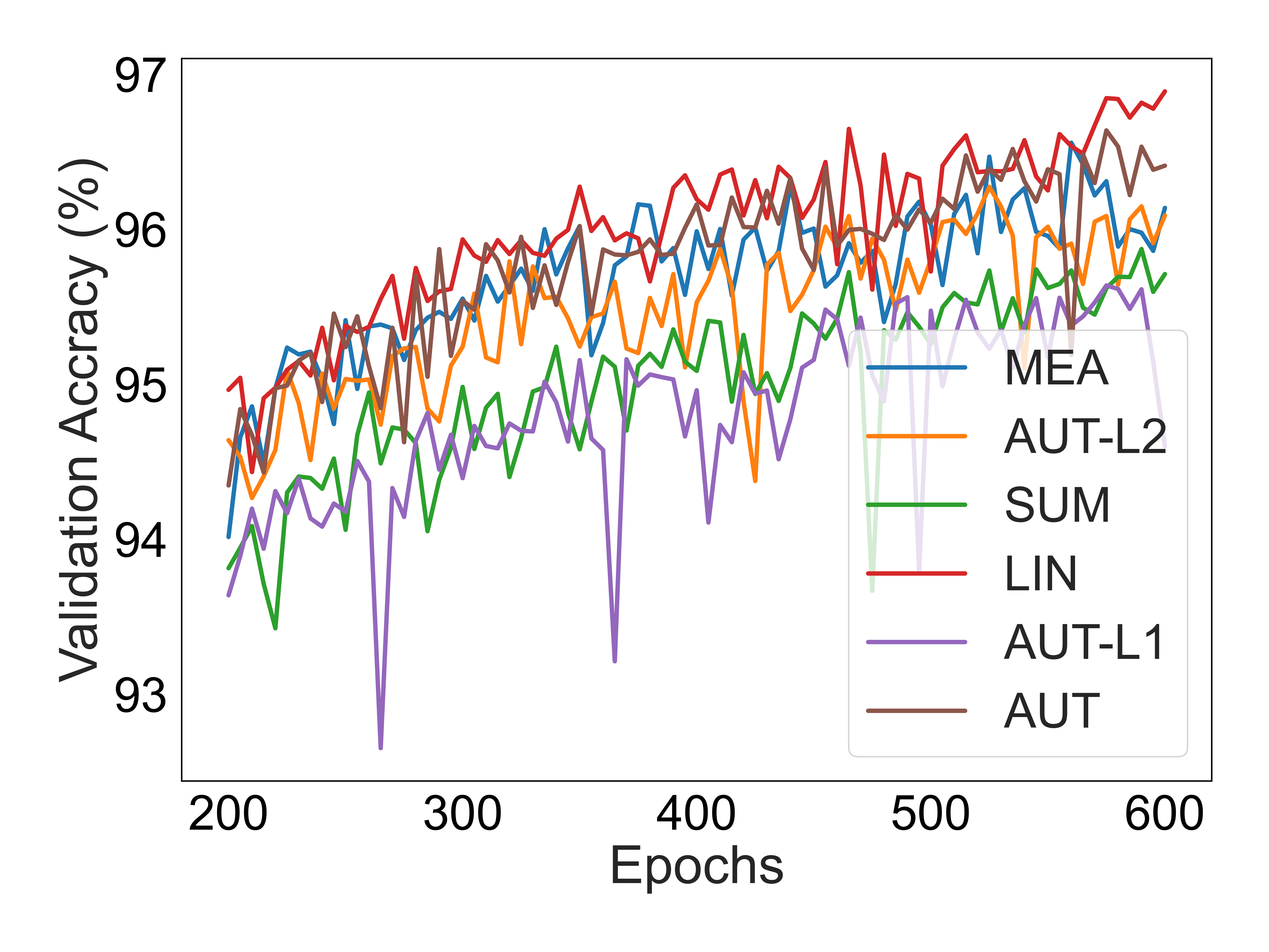}
    \caption{The performance analysis of various operators in DRNet on the RAVEN dataset reveals that the $\mathbf{LIN}$ operator outperforms its counterparts. }
    \label{figure: trn loss and tst acc}
\end{figure}

In the last approach, we concatenate these two streams and implement the learnable attention operator $\mathbf{LIN}$ using a linear layer:
\begin{equation}
    \mathbf{LIN}(\cdot) := concat(x^{cnn\_out}, x^{vit\_out}) A ^{T} + b
\end{equation}
We compare different operators in Figure \ref{figure: trn loss and tst acc} and adopt the $\mathbf{LIN}$ operator in DRNet.

After feature fusion, we split the fused features $x$ into two groups: $e_{i}$ and $c_{i}$, where $i \in {1, 2, ..., 8}$. We then concatenate each $c_{i}$ with the 8 context features to form $r_{i} = [e_{1}, e_{2}, ..., e_{8}, c_{i}]$, with $r_{i}\in\mathcal{R}^{(B, i, 9, 400)}$. Next, we pass $r_{i}$ into the rule extractor to infer the relationships between the nine feature vectors, as depicted in the reasoning module of Figure \ref{fig:framework}.

\textbf{Rule Extractor.}
The rule extractor consists of two ResBlocks. Each residual branch has two 1D convolutional layers with a kernel size of 7. Each convolutional layer learns to expand the receptive fields of the neurons to extract higher-level relations with a stride of 1. The shortcut connection applies a 1D convolutional layer to the two ResBlocks with a kernel size and stride of 1. In total, our rule extractor can be formulated as:
\begin{equation}
    r_{i}^{l} = ReLU(BN(Conv_{7}(r_{i}^{l-1}))), l\in{1,2}
\end{equation}
where $l$ represents the layer index of convolutional layers. We set the filters to $[64, 128]$ for the first and second convolutional layers. After the skip connection, we apply $MaxPool1d(r_{i}^{1})$ to reshape $r_{i}\in\mathcal{R}^{(B, i, 128, 400)}$ into $r_{i}\in\mathcal{R}^{(B, i, 128, 100)}$.
Then, we send $r_{i}$ to the second ResBlocks as follows:
\begin{equation}
    r_{i}^{l} = ReLU(BN(Conv_{7}(r_{i}^{l-1}))), l\in{3,4}
\end{equation}
We set the filters to $[128, 64]$ for the third and fourth layers. After the skip connection, we apply $AdaptiveAvgPool1d(r_{i}^{1})$ to reshape $r_{i}\in\mathcal{R}^{(B, i, 64, 100)}$ into $r_{i}\in\mathcal{R}^{(B, i, 64, 16)}$. Finally, we flatten $r_{i}\in\mathcal{R}^{(B, i, 64, 16)}$ into $r_{i}\in\mathcal{R}^{(B, i, 1024)}$ to obtain 8 embeddings. The embeddings corresponding to the correct labels are both abstract representations of the rules.

\textbf{Classifier.} Lastly, we use an $\mathbf{MLP}$ consisting of three linear layers to score these features, and the highest score determines the best answer:
\begin{equation}
    Answer = \mathop{\arg\max}_{i \in \{1, ..., 8\}} [\mathbf{MLP}(r_{i})]
\end{equation}
Between every two linear layers, we have added an $ELU$ function and a $BatchNorm1d$ layer, with a dropout probability of 0.5. For each linear layer, the output dimensions are 512, 256, and 1 respectively.

\section{Experiments}
\subsection{Datasets}
The \textbf{PGM} dataset \cite{santoro2018measuring} comprises 1.2 million training samples, 20 thousand validation samples, and 200 thousand test samples. Each panel within the PGM dataset varies in terms of types, sizes, colors, and shapes. There are 1-4 rules per row or column for each matrix panel. The PGM dataset includes 8 regimes, with 7 of them involving interpolation, extrapolation, held-out attribute pairs (HO AP), held-out pairs of triples (HO TP), held-out triples (HO Triples), held-out line-type (HO LT), and held-out shape-color (HO SC) scenarios. These regimes systematically assess out-of-distribution (OOD) generalization using various approaches.

\textbf{RAVEN-style Datasets.} The RAVEN dataset \cite{zhang2019raven}, along with its variants I-RAVEN \cite{hu2020stratified} and RAVEN-FAIR \cite{benny2021scale}, are compact datasets, each comprising 7 configurations, with each configuration containing 10,000 samples.
The distribution ratio across the training, test, and validation sets is 6:2:2. Every problem within these datasets features 4-8 rules in each row/column. Each problem is presented with 8 context panels arranged in an incomplete 3 x 3 matrix, alongside 8 candidate answer panels. We trained DRNet jointly on all configurations in each RAVEN-style dataset.

\subsection{Implementation Details}
All datasets include training, validation, and test sets. We utilize a standard batch size of 256 and evaluate the reported accuracy on the test set using the best validation accuracy checkpoint. The same set of hyperparameters is applied across all benchmarks, employing Adam \cite{da2014method} optimizer with a learning rate of 3e-4, $\beta$ values of (0.9, 0.999), and a weight decay of 1e-6. No additional supervision signals, such as metadata, are utilized during training. Additionally, for RAVEN-style datasets, we present the median outcome from 5 distinct runs. Given the computational demands of training on large-scale PGM datasets, we provide a single result, aligning with the approach of prior works \cite{zhang2019learning, benny2021scale, pmlr-v202-yang23r}. 
In the experimental results presented below, when the validation loss no longer decreases within 20 epochs, we perform early stopping.
\begin{table*}[h]
	\centering
	{\begin{tabular}{l|cccccccccc|c}
			\hline
			Method & WReN & CoPINet & MRNet & SCL                                  & MLRN                     & Rel-Base & ARII            & STSN & PredRNet & NVSA  & DRNet\\
			\hline
			PGM-N& 62.6     &  56.4        &94.5      & 88.9                                  &98.0                          &85.5         &88.0                &98.2 &97.4 &-       &\textbf{99.06}      \\
			RAVEN & 16.8     &  91.4         &96.6      & 91.6                                  & 12.3$^\dagger$   &91.7             &-                   & 89.7$^\dagger$  &95.8       &87.7 &\textbf{96.89}   \\
			I-RAVE &   23.8   &  46.1         &83.5$^\dagger$      & 95.0          &  12.3$^\dagger$  &91.1$^\dagger$   &91.1   &95.7$^\dagger$  &96.5 &88.1 &\textbf{97.62}   \\
			RAVE-F &  30.3  &  50.6         &88.4      & 90.1$^\dagger$           & 29.5$^\dagger$  &93.5$^\dagger$ &-           &95.4  &97.1        &-       &\textbf{97.58}\\
			\hline
			Average&33.4 &61.1&90.8&91.4&38.0&90.5&-&94.8&96.7&-&\textbf{97.79} \\
			\hline
	\end{tabular}}
	\caption{Recognition accuracy ($\%$) on PGM Neutral (PGM-N), RAVEN, I-RAVEN (I-RAVE), and RAVEN-FAIR (RAVE-F). For all RAVEN-style datasets, accuracy is obtained by averaging across all seven configurations. $^\dagger$ indicates that the results were not reported in the original paper; we obtained these results from Table 1 of \cite{pmlr-v202-yang23r}. The best results for each dataset are highlighted in bold font.}
	\label{table:summary performance}
\end{table*}
\subsection{Main Results}
We conducted experiments on PGM, RAVEN, I-RAVEN and RAVEN-FAIR, all of which have predefined training, validation and test data splits. During training, we used the training set for model training and the test set for evaluation, while the validation set was used to select the optimal checkpoint for evaluation. We used vertical/horizontal flip data augmentation with a probability of 0.3 for RPM training samples.

\textbf{State-of-the-art Comparisons.}
We compare DRNet with several previous models, including WReN \cite{santoro2018measuring}, CoPINet \cite{zhang2019learning}, MRNet \cite{benny2021scale}, SCL \cite{wu2020scattering}, MLRN \cite{jahrens2020solving}, Rel-Base \cite{spratley2020closer}, ARII \cite{zhang2022learning}, STSN \cite{mondallearning} and PredRNet \cite{pmlr-v202-yang23r}. We have also compared our method with end-to-end symbolic methods such as NVSA~\cite{hersche2023neuro}. Experiments were conducted on PGM \textit{Neutral} and three RAVEN-style datasets.

Table \ref{table:summary performance} shows the main results on four datasets.
First, our DRNet achieves the best average performance on the four datasets compared to single-stream models, such as STSN and PredRNet. STSN introduces slot attention to extract image-wise features and then proposes a transformer-based module to explore relationships between contexts and choices for reasoning. PredRNet introduces prediction error into ConvBlocks to improve reasoning performance. PredRNet provides the best average performance (96.7\%) among all compared methods. While DRNet outperforms PredRNet with an average performance of 97.78\%.
In addition, DRNet actually achieves better performance on the RAVEN (1.09\%), I-RAVEN (+1.12\%), and RAVEN-FAIR (+0.48\%) datasets than PredRNet, respectively.
Some recently proposed methods, such as STSN, MLRN, and ARII, only show good results on one or two datasets. For example, CoPINet and MLRN only perform well on RAVEN (91.4\%) and PGM-N (98.0\%) respectively.
In contrast, DRNet shows superior results on all 4 datasets, which suggests the generalization performance on different datasets.
Second, compared to certain competitive models like MRet, our approach extracts rules only from combinations of 8 image sets, without the need for column rule learning.
If we remove the ViT branch in DRNet, our model is similar to Rel-Base. If we remove the CNN branch, our model completely degenerates into an attention-based model. However, neither branch performed well enough (see ablation experiment), indicating that the two-stream design of our model was critical and helped our model achieve an average result of 97.78\%.

\textbf{Out-of-Distribution Generalization in PGM.} 
In addition to the Neutral dataset of PGM, the remaining seven datasets were employed to evaluate the model's capacity to handle out-of-distribution scenarios. We assessed DRNet's performance across all sub-datasets of PGM while maintaining consistent model settings. The outcomes of these evaluations are meticulously documented in Table \ref{table:VRNet on pgm}. A careful examination of the data presented in Table 1 reveals that our proposed model exhibited only a marginal average enhancement of 1.09\% for in-distribution datasets. However, when confronting challenges posed in out-of-distribution scenarios, our model showcased a notable enhancement, reaching up to 11.23\%. This highlights the robust learning prowess inherent in the dual-stream architecture, enabling it to effectively handle OOD challenges.
\begin{table*}[t]
	\centering
	\begin{tabular}{l|cccccccc|c}
		\hline
		Method    & Neut         & Intr & Extr & HO AP & HO TP& HO Tri &HO LT &HO SC & Average \\
		\hline
		WReN      & 62.6 &62.4 &17.2 &27.2 &41.9 &19.0 &14.4 &12.5 &32.4\\
		ARII           &88.0 &72.0 &\textbf{29.0} &50.0 &64.1 &32.1 &16.0 &12.7 &45.49\\
		MXGNet  &66.7 &65.4  &18.9 &33.6 &43.3 &19.9 &16.7 &16.6 &35.14\\
		MRNet     &93.4 &68.1  &19.2 &38.4 &55.3 &25.9 &\textbf{30.1} &\textbf{16.9}  &43.41\\
		PredRNet&97.4 &70.5 &19.7 &63.4 &67.8 &23.4 &27.3 &13.1&47.1\\
		\hline
		DRNet     &\textbf{99.06} &\textbf{83.78} &22.22 &\textbf{93.74} &\textbf{78.11} &\textbf{48.77} &27.92 &13.09 &\textbf{58.33}\\
		\hline
	\end{tabular}
	\caption{Recognition accuracy (\%) across all regimes of PGM. PGM comprises 1 Neutral and 7 OOD sub-datasets. Neut: Neutral, Inter: Interpolation; Extr: Extrapolation; HP AP: Held-Out Attribute Pairs; HO TP: Held-Out Triple Pairs; HO Tri: Held-Out Triples; HO LT: Held-Out Line Type; HO SC: Held-Out Shape Color. The best results are highlighted in bold font.}
	\label{table:VRNet on pgm}
\end{table*}
\subsection{Ablation Experiments}
We conducted ablation experiments on both the I-RAVEN dataset, representing in-distribution, and the PGM HO AP dataset, representing OOD. Since the three RAVEN-style datasets have similar distributions, we only tested on one.

\textbf{Different hyper-parameters.}
We evaluated the impact of different depths of ViTs and various sizes of convolutional kernels on model performance. We selected ViT depths of 4, 8, 12 (DRNet), and 16. Regarding convolution, we investigated common kernel sizes: 3, 5, and 7 (DRNet). The results are presented in Table \ref{ablation_different_hyper_parameters}. DRNet demonstrates a gradual improvement in performance with increasing ViT depth. 
However, when the depth is 16, the performance of DRNet on HO AP decreases from 93.7\% to 89.7\%. Additionally, DRNet's performance gradually improves with larger convolutional kernel sizes on both datasets.  These results indicate that the current hyperparameters for DRNet are optimal.

\textbf{Single stream \textit{v.s.} Dual stream.}
Compared with single-stream models like MRNet and PredRNet, DRNet performs very well, especially in the OOD scenario. To help understand our proposed model, DRNet, the first thing we aimed to clarify is the role of each branch. The results are shown in Table \ref{ablation_single_vs_dual}.

Each stream achieves a recognition accuracy of over 80\% on the in-distribution I-RAVEN dataset. 
\begin{table}[htbp]
	\centering
	\begin{tabular}{@{\hspace{0.5em}}c@{\hspace{0.8em}}c@{\hspace{0.8em}}c@{\hspace{0.8em}}c@{\hspace{0.8em}}c|@{\hspace{1em}}c@{\hspace{1em}}c@{\hspace{1em}}c}
		\hline
		& \multicolumn{4}{c}{ViT Depth}&\multicolumn{3}{c}{Conv kernel}\\
		\hline
		Dataset & 4 & 8& \textbf{12} & 16& 3 & 5 &\textbf{7}\\
		\hline
		I-RAVE                           & 96.4 & 96.0               &\textbf{97.6} &96.2 &90.9  &95.8&\textbf{97.6}              \\
		HO AP                           &   93.4     & 93.4          &\textbf{93.7}  &89.7 &84.6  &91.5&\textbf{93.7}  \\
		\hline
	\end{tabular}
	\caption{Recognition accuracy (\%) of different hyper-parameters on I-RAVE and PGM HO AP.  The results, in bold, show that the hyper-parameters used in DRNet are optimal. I-RAVE represents I-RAVEN; HO AP corresponds to PGM Held-Out Attribute Pairs.}
	\label{ablation_different_hyper_parameters}
\end{table}
Data augmentation (DA) helps the model achieve higher performance on 
such datasets; a single CNN stream with data augmentation achieves a recognition accuracy of 95.50\%, surpassing many previous studies. The contribution of the added ViT stream to the performance improvement in DRNet appears to be low.
 Hence, we also conducted an ablation experiment on the OOD PGM HO AP dataset. As shown in Table \ref{ablation_single_vs_dual}, the recognition accuracy for ViT+DA is 72.26\%, while for CNN+DA, it is only 62.87\%. The performance of the single stream is far inferior to the dual-stream architecture of DRNet, indicating that the dual-stream network enhances the model's performance on OOD datasets.

\begin{table}[t]
	\centering
	\begin{tabular}{ccc|cc}
		\hline
		ViT & CNN & DA & I-RAVE & HO AP \\
		\hline
		$\surd$ & $\times$& $\times$ &83.62&67.53\\
		$\surd$  & $\times$& $\surd$  &87.52&72.26\\
		$\times$ & $\surd$ & $\times$ &90.83&58.73\\
		$\times$ & $\surd$ & $\surd$ &95.50&62.87\\
		$\surd$  & $\surd$ & $\times$ & 91.68&90.46\\
		$\surd$  & $\surd$ & $\surd$ &\textbf{97.62}&\textbf{93.74}\\
		\hline
	\end{tabular}
	\caption{Recognition accuracy (\%) of different branches across various datasets. CNN refers to the CNN branch mentioned earlier; DA stands for Data Augmentation, and we used vertical/horizontal flip data augmentation with a probability of 0.3 for RPM training samples.}
	\label{ablation_single_vs_dual}
\end{table}

\textbf{Dual CNN Stream \textit{v.s.} Dual ViT Stream.} 
Subsequently, we embarked on the replacement of the network's branches to ascertain whether any dual-stream model comprising distinct network components possesses adept relational reasoning capabilities. Given that the tensor shapes of input and output for each stream in DRNet remain consistent, our focus is centered on assessing whether the parameter count of the new networks was comparable to that of DRNet. 
\begin{table}[ht]
	\centering
	\begin{tabular}{l|cc}
		\hline
		Method & I-RAVE & HO AP \\
		\hline
		DCNet(24M)&\textbf{98.24}& 73.78\\
		DVNet-s(26M) &87.52& 92.60\\
		DVNet-h(47M)&97.32&93.24\\
		DRNet-P(3.4M)&96.06&91.23\\
		DRNet(24.6M)&97.62&\textbf{93.74}\\
		\hline
	\end{tabular}
	\caption{Recognition accuracy (\%) of different dual-stream networks on I-RAVEN and PGM HO AP datasets, where DCNet refers to the Dual-CNN network, DVNet refers to the Dual-ViT network, and DRNet-P refers to the patch-based DRNet.}
	\label{ablation_dual_comparision}
\end{table}
For simplicity of description, we define the CNN branch network in DRNet as CNN-base and the ViT branch as ViT-base.

First, we replace the ViT-base in DRNet with ResNet-32 to form a dual-CNN network (DCNet-24M), where M represents learnable parameters, in millions. For the ResBlocks, we set filters to $[64\times3, 128\times4, 256\times6, 512\times3]$, where $[3, 4, 6, 3]$ represents the repeat times of ResBlocks, and all convolutional layers in the ResBlocks have a kernel size of 7. The first convolutional layer in ResNet-32 has a kernel size of 3, and the filter is set to 64.

Second, we replace the CNN-base in DRNet with a shallow ViT-small of depth 1 to create a dual-ViT network (DVNet-s-26M). The network architecture of ViT-small is the same as ViT-base.

Third, we use two ViT-b models to construct a larger dual-ViT network (DVNet-h-47M). 
Fourth, DVNet-h increases the number of parameters drastically. Given that we are applying attention to the sum of ViT-base and CNN-base, we instead utilized a patch-based CNN from \cite{brutzkus2022efficient} , replacing ViT-base to create a patch-based DRNet (DRNet-P-3.4M).
We tested the four aforementioned network configurations on the I-RAVEN and PGM HO AP datasets, and the results are shown in Table \ref{ablation_dual_comparision}.

From Table \ref{ablation_dual_comparision}, it can be observed that DCNet performs excellently on the I-RAVEN dataset (98.24\%), surpassing DRNet (97.62\%). However, DCNet's performance drastically declines on the OOD PGM HO AP dataset, achieving only 73.78\%.
Nevertheless, DCNet's results still outperform those of previous state-of-the-art models, such as PredRNet.
In contrast, DVNet-s performs well in both the I-RAVEN and OOD paradigms, indicating that spatial attentional representations at different scales are crucial for the abstract reasoning process.
As the number of learnable parameters in DVNet-h increases, its performance aligns with that of DRNet, but at a higher computational cost. This suggests that DRNet strikes an effective balance between computational complexity and performance. 
Compared to other baseline models, DRNet has a larger number of parameters. Therefore, we introduced a small-parameter version called DRNet-P. As shown in Table \ref{ablation_dual_comparision}, DRNet-P exhibits a significant reduction in parameters by 86.2\% (24.6M$\rightarrow$3.4M) compared to DRNet. 
Despite this reduction, its performance only slightly decreased by 1.56\% on the I-RAVE dataset and by 2.51\% on the PGM HO AP dataset. 
This indicates the potential of dual-stream networks in abstract visual reasoning.

\subsection{Rule Representations}
\begin{figure}[ht]
	\centering
	\includegraphics[scale=0.53]{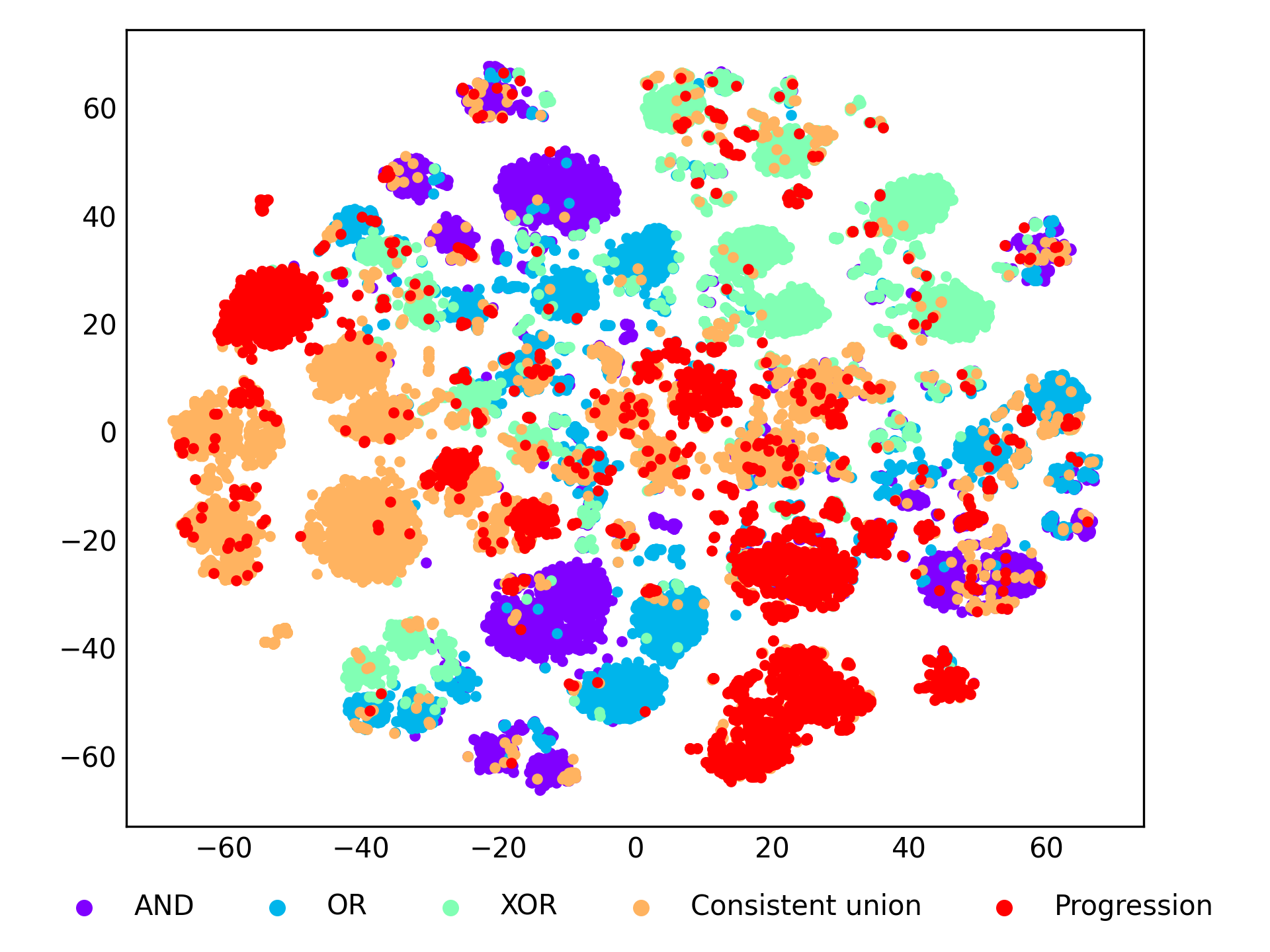}
	\caption{T-SNE visualization of abstract rules in PGM Neutral. The clustered embedding of similar abstract rules indicates that the Reasoning Module in DRNet is adept at discovering rules.
	} 
	\label{fig:tsne_visualization.png}
\end{figure}

Although DRNet achieves a high recognition accuracy, we still lack an understanding of its reasoning process. All rule extraction takes place within the Rule Extractor.
Therefore, we conducted a t-SNE \cite{tsne:vandermaaten08a} analysis of the embeddings corresponding to the correct answers of the rule extractor.
Due to the variety of rules covered by RAVEN problems, it is challenging to visualize vectors with multiple rule labels. Therefore, we selected the PGM Neutral dataset for rule visualization due to its smaller number of rules. Based on the actual rule types of each problem (AND, OR, XOR, Consistent union, Progression), we visualized 200k test samples, and the results are shown in Figure \ref{fig:tsne_visualization.png}. 
The rule extractor module can identify and form abstract rule representations for downstream classification task. 

Furthermore, we utilized radial basis function kernel principal component analysis  to reduce the 200k rule representations from 1024 to 768 dimensions, preserving spatial information and reducing redundancy. Subsequently, we computed pairwise cosine similarities for the 5 rule categories, resulting in 10 sets of scores. 
The mean values of these scores range from -0.01 to 0.007, indicating near orthogonal representations and this may be a reason for the superior performance of DRNet.

\subsection{Visualization of Two Streams}
\citeauthor{cadieu2012learning} shows that learning both ventral and dorsal-like representations in a single ANN with two pathways is possible if one forces the two pathways to process separately the phase and amplitude of a complex decomposition of the stimuli \cite{cadieu2012learning}. 

\begin{figure}[ht]
	\centering
	\includegraphics[scale=0.52]{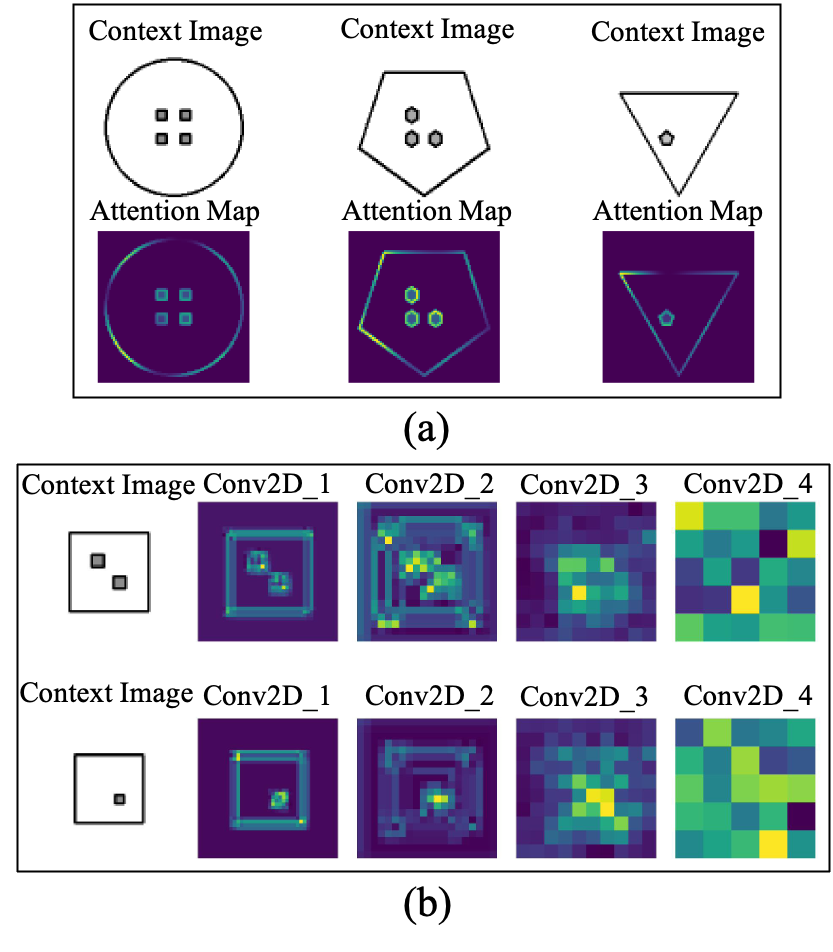}
	\caption{Illustrative examples from ViT and CNN streams. (a) Demonstrative instances of attention mapping from output tokens to the input space. (b) Visualization of the convolutional layer obtained through the forward process of DRNet.
	} 
	\label{fig:visual_fo_vit_and_cnn}
\end{figure}

To begin to understand how the two streams in DRNet process images, we analyze their internal representations. Self-attention allows ViT to integrate information across the entire image even in the lowest layers. We investigate to what degree the network makes use of this capability. For the analysis of ViT, we adopt the approach used in \cite{vaswani2017attention}. We find that ViT attends to image regions that are relevant for spatial information, as shown in Figure \ref{fig:visual_fo_vit_and_cnn} (a), working like the \textit{where} pathway.

For the visualization of CNN convolutional layers, we obtained the results through a single forward pass, as shown in Figure \ref{fig:visual_fo_vit_and_cnn} (b). We found that during the learning process, CNN gradually acquires high-level image representations by combining local features, working like the \textit{what} pathway.

\begin{figure}[htbp]
	\centering
	\includegraphics[scale=0.22]{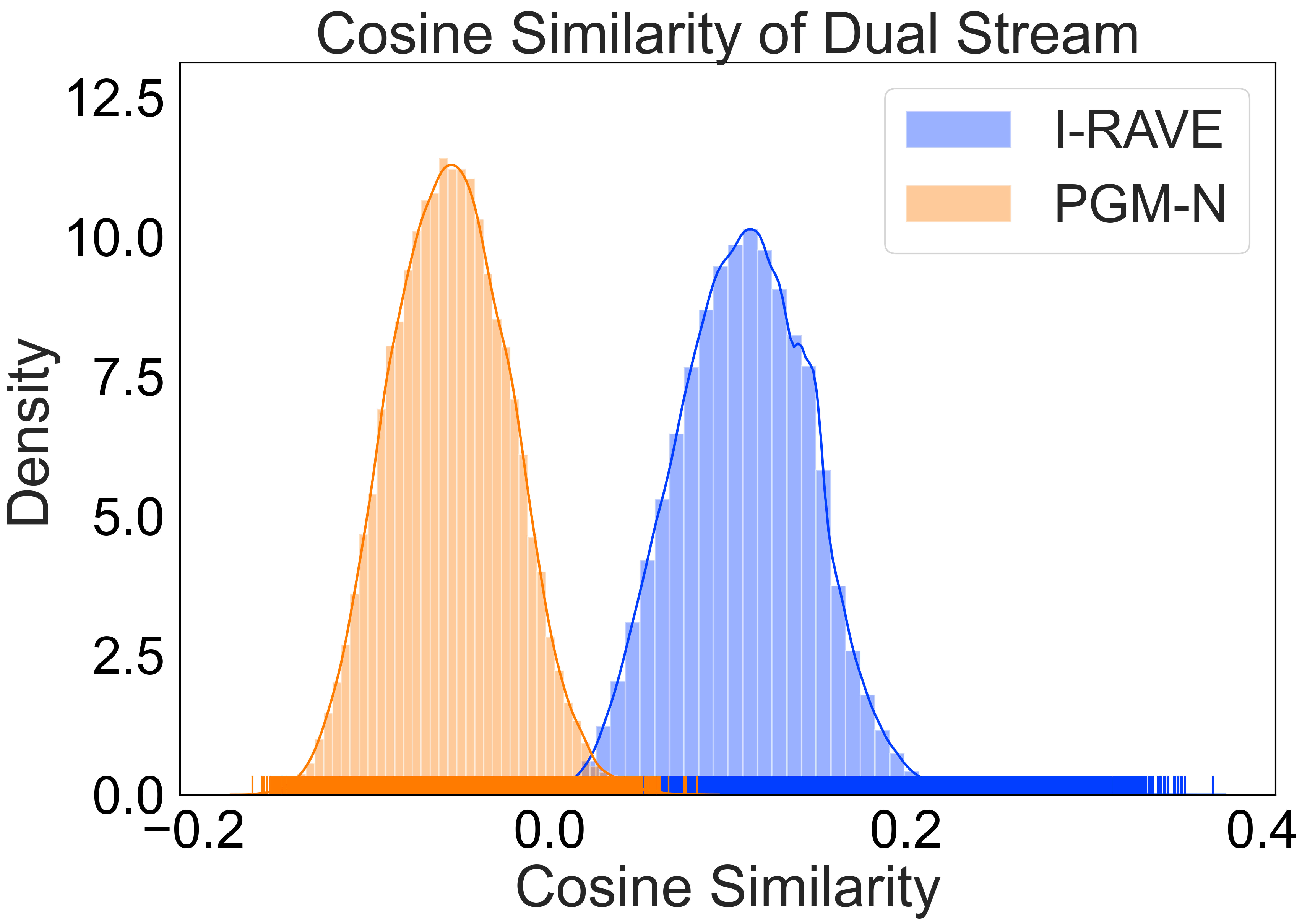}
	\caption{Cosine similarity for dual encoder representations. We calculated the cosine similarity between dual encoder representations for each test sample and displayed the statistical data in the form of a histogram and a rug plot.} 
	\label{fig:cosine_similarity_vit_and_cnn}
\end{figure}

We also investigated the similarity between the emerging representations. We computed the cosine similarity for dual encoder representations on PGM-N and I-RAVE test sets (Figure \ref{fig:cosine_similarity_vit_and_cnn}). It can be seen that in both datasets, the learned dual encoder representations exhibit small cosine similarities.
\section{Discussion}
DRNet shows superior performance on multiple datasets, highlighting the potential of this dual-stream architecture. In our work, we use two backbone networks to mimic the two streams of visual processing in mammalian brains, and the high-level features obtained by these two streams form clearer discrete abstract rules through a rule extractor.
With its remarkable generalization performance and accuracy across multiple benchmarks, DRNet provides an effective and powerful baseline in visual abstract reasoning.

To our knowledge, there is no existing research on dual-stream architectures within these benchmarks. To encourage further work, the technical shortcomings of the model are highlighted and explained in detail. 
The first question is whether a larger convolution kernel affects the model's ability to extract features,
 and the second is that the visual transformer module we used selected a large patch size to match the size of the CNN branch, which may have affected the visual transformer's ability to generalize global and local information.
Currently, our model has poor recognition results in RAVEN-style $3 \times 3$ grid configuration. This may be due to the size of the convolution kernel and the patch size of ViT, which needs further investigation in the future. 

The two-stream hypothesis involves many brain regions, and rigorous modeling is very challenging. At the framework level, we can extend our model to include more regions, in particular the hippocampal formation, which has long been considered as the basis for memory formation, flexible decision-making and reasoning \cite{behrens2018cognitive, whittington2020tolman, whittington2022build}. In \cite{bakhtiari2021functional}, the functional specialization of the visual cortex emerges from training parallel pathways with self-supervised predictive learning; the potential to incorporate such functionality into DRNet for similar tasks is still under exploration.

Multimodal perception is necessary to achieve general artificial intelligence. Models combined with language models can exhibit zero-shot or few-shot learning capabilities in such non-verbal reasoning tasks \cite{huang2023language}, and our framework can flexibly integrate multimodal information in the future.

\section{Conclusion}

We applied DRNet to multiple datasets and observed that it achieves a high level of recognition accuracy and demonstrates an ability to generalize. Our experiments revealed that the dual-stream framework does not lead to the superposition of effects from the respective branches. In ablation experiments, we discovered that having more learnable parameters in a dual-stream network does not necessarily result in improved performance, and the rule extractor designed in DRNet can learn discrete abstract rule representations. We have illustrated the effectiveness of this work and its potential for abstract visual reasoning.

\section{Acknowledgments}
This work is partly supported by National Key R\&D Program of China (2019YFA0709503). We thank Xiaohong Wan, Dahui Wang, Hongzhi You, Ruyuan Zhang and Zonglei Zhen for fruitful discussions.
\bibliography{aaai24}

\begin{thebibliography}{61}
\providecommand{\natexlab}[1]{#1}

\bibitem[{Bakhtiari et~al.(2021)Bakhtiari, Mineault, Lillicrap, Pack, and
  Richards}]{bakhtiari2021functional}
Bakhtiari, S.; Mineault, P.; Lillicrap, T.; Pack, C.; and Richards, B. 2021.
\newblock The functional specialization of visual cortex emerges from training
  parallel pathways with self-supervised predictive learning.
\newblock \emph{Advances in Neural Information Processing Systems}, 34:
  25164--25178.

\bibitem[{Behrens et~al.(2018)Behrens, Muller, Whittington, Mark, Baram,
  Stachenfeld, and Kurth-Nelson}]{behrens2018cognitive}
Behrens, T.~E.; Muller, T.~H.; Whittington, J.~C.; Mark, S.; Baram, A.~B.;
  Stachenfeld, K.~L.; and Kurth-Nelson, Z. 2018.
\newblock What is a cognitive map? Organizing knowledge for flexible behavior.
\newblock \emph{Neuron}, 100(2): 490--509.

\bibitem[{Benny, Pekar, and Wolf(2021)}]{benny2021scale}
Benny, Y.; Pekar, N.; and Wolf, L. 2021.
\newblock Scale-localized abstract reasoning.
\newblock In \emph{Proceedings of the IEEE/CVF Conference on Computer Vision
  and Pattern Recognition}, 12557--12565.

\bibitem[{Brutzkus et~al.(2022)Brutzkus, Globerson, Malach, Netser, and
  Shalev-Schwartz}]{brutzkus2022efficient}
Brutzkus, A.; Globerson, A.; Malach, E.; Netser, A.~R.; and Shalev-Schwartz, S.
  2022.
\newblock Efficient Learning of CNNs using Patch Based Features.
\newblock In \emph{International Conference on Machine Learning}, 2336--2356.
  PMLR.

\bibitem[{Cadieu and Olshausen(2012)}]{cadieu2012learning}
Cadieu, C.~F.; and Olshausen, B.~A. 2012.
\newblock Learning intermediate-level representations of form and motion from
  natural movies.
\newblock \emph{Neural computation}, 24(4): 827--866.

\bibitem[{Carreira and Zisserman(2017)}]{carreira2017quo}
Carreira, J.; and Zisserman, A. 2017.
\newblock Quo vadis, action recognition? a new model and the kinetics dataset.
\newblock In \emph{proceedings of the IEEE Conference on Computer Vision and
  Pattern Recognition}, 6299--6308.

\bibitem[{Chen et~al.(2022)Chen, Zuo, Wei, Wu, Liu, and Mak}]{chen2022two}
Chen, Y.; Zuo, R.; Wei, F.; Wu, Y.; Liu, S.; and Mak, B. 2022.
\newblock Two-stream network for sign language recognition and translation.
\newblock \emph{Advances in Neural Information Processing Systems}, 35:
  17043--17056.

\bibitem[{Court(1982)}]{Court1982manual}
Court, J. 1982.
\newblock Manual for Raven’s Progressive Matrices and Vocabulary Scales.

\bibitem[{Cui, Liu, and Zhang(2019)}]{cui2019deep}
Cui, R.; Liu, H.; and Zhang, C. 2019.
\newblock A deep neural framework for continuous sign language recognition by
  iterative training.
\newblock \emph{IEEE Transactions on Multimedia}, 21(7): 1880--1891.

\bibitem[{Da(2014)}]{da2014method}
Da, K. 2014.
\newblock A method for stochastic optimization.
\newblock \emph{arXiv preprint arXiv:1412.6980}.

\bibitem[{Dosovitskiy et~al.(2021)Dosovitskiy, Beyer, Kolesnikov, Weissenborn,
  Zhai, Unterthiner, Dehghani, Minderer, Heigold, Gelly, Uszkoreit, and
  Houlsby}]{dosovitskiy2020image}
Dosovitskiy, A.; Beyer, L.; Kolesnikov, A.; Weissenborn, D.; Zhai, X.;
  Unterthiner, T.; Dehghani, M.; Minderer, M.; Heigold, G.; Gelly, S.;
  Uszkoreit, J.; and Houlsby, N. 2021.
\newblock An Image is Worth 16x16 Words: Transformers for Image Recognition at
  Scale.
\newblock In \emph{9th International Conference on Learning Representations,
  {ICLR} 2021, Virtual Event, Austria, May 3-7, 2021}. OpenReview.net.

\bibitem[{Feichtenhofer et~al.(2019)Feichtenhofer, Fan, Malik, and
  He}]{feichtenhofer2019slowfast}
Feichtenhofer, C.; Fan, H.; Malik, J.; and He, K. 2019.
\newblock Slowfast networks for video recognition.
\newblock In \emph{Proceedings of the IEEE/CVF international conference on
  computer vision}, 6202--6211.

\bibitem[{Feichtenhofer, Pinz, and
  Zisserman(2016)}]{feichtenhofer2016convolutional}
Feichtenhofer, C.; Pinz, A.; and Zisserman, A. 2016.
\newblock Convolutional two-stream network fusion for video action recognition.
\newblock In \emph{Proceedings of the IEEE conference on computer vision and
  pattern recognition}, 1933--1941.

\bibitem[{Girshick et~al.(2015)Girshick, Donahue, Darrell, and
  Malik}]{girshick2015region}
Girshick, R.; Donahue, J.; Darrell, T.; and Malik, J. 2015.
\newblock Region-based convolutional networks for accurate object detection and
  segmentation.
\newblock \emph{IEEE transactions on pattern analysis and machine
  intelligence}, 38(1): 142--158.

\bibitem[{Goodale and Milner(1992)}]{goodale1992separate}
Goodale, M.~A.; and Milner, A.~D. 1992.
\newblock Separate visual pathways for perception and action.
\newblock \emph{Trends in neurosciences}, 15(1): 20--25.

\bibitem[{Grezes and Decety(2002)}]{grezes2002does}
Grezes, J.; and Decety, J. 2002.
\newblock Does visual perception of object afford action? Evidence from a
  neuroimaging study.
\newblock \emph{Neuropsychologia}, 40(2): 212--222.

\bibitem[{Hahne et~al.(2019)Hahne, L{\"u}ddecke, W{\"o}rg{\"o}tter, and
  Kappel}]{hahne2019attention}
Hahne, L.; L{\"u}ddecke, T.; W{\"o}rg{\"o}tter, F.; and Kappel, D. 2019.
\newblock Attention on abstract visual reasoning.
\newblock \emph{arXiv preprint arXiv:1911.05990}.

\bibitem[{He et~al.(2016)He, Zhang, Ren, and Sun}]{he2016deep}
He, K.; Zhang, X.; Ren, S.; and Sun, J. 2016.
\newblock Deep residual learning for image recognition.
\newblock In \emph{Proceedings of the IEEE conference on computer vision and
  pattern recognition}, 770--778.

\bibitem[{Hersche et~al.(2023)Hersche, Zeqiri, Benini, Sebastian, and
  Rahimi}]{hersche2023neuro}
Hersche, M.; Zeqiri, M.; Benini, L.; Sebastian, A.; and Rahimi, A. 2023.
\newblock A neuro-vector-symbolic architecture for solving Raven’s
  progressive matrices.
\newblock \emph{Nature Machine Intelligence}, 1--13.

\bibitem[{Hoshen and Werman(2017)}]{hoshen2017iq}
Hoshen, D.; and Werman, M. 2017.
\newblock Iq of neural networks.
\newblock \emph{arXiv preprint arXiv:1710.01692}.

\bibitem[{Hu et~al.(2021)Hu, Ma, Liu, Wei, and Bai}]{hu2020stratified}
Hu, S.; Ma, Y.; Liu, X.; Wei, Y.; and Bai, S. 2021.
\newblock Stratified Rule-Aware Network for Abstract Visual Reasoning.
\newblock In \emph{Thirty-Fifth {AAAI} Conference on Artificial Intelligence,
  {AAAI} 2021, Thirty-Third Conference on Innovative Applications of Artificial
  Intelligence, {IAAI} 2021, The Eleventh Symposium on Educational Advances in
  Artificial Intelligence, {EAAI} 2021, Virtual Event, February 2-9, 2021},
  1567--1574. {AAAI} Press.

\bibitem[{Huang et~al.(2021)Huang, Rolls, Hsu, Feng, and
  Lin}]{huang2021extensive}
Huang, C.-C.; Rolls, E.~T.; Hsu, C.-C.~H.; Feng, J.; and Lin, C.-P. 2021.
\newblock Extensive cortical connectivity of the human hippocampal memory
  system: beyond the “what” and “where” dual stream model.
\newblock \emph{Cerebral Cortex}, 31(10): 4652--4669.

\bibitem[{Huang et~al.(2023)Huang, Dong, Wang, Hao, Singhal, Ma, Lv, Cui,
  Mohammed, Patra, Liu, Aggarwal, Chi, Bjorck, Chaudhary, Som, Song, and
  Wei}]{huang2023language}
Huang, S.; Dong, L.; Wang, W.; Hao, Y.; Singhal, S.; Ma, S.; Lv, T.; Cui, L.;
  Mohammed, O.~K.; Patra, B.; Liu, Q.; Aggarwal, K.; Chi, Z.; Bjorck, J.;
  Chaudhary, V.; Som, S.; Song, X.; and Wei, F. 2023.
\newblock Language Is Not All You Need: Aligning Perception with Language
  Models.
\newblock \emph{CoRR}, abs/2302.14045.

\bibitem[{Jaeggi et~al.(2008)Jaeggi, Buschkuehl, Jonides, and
  Perrig}]{jaeggi2008improving}
Jaeggi, S.~M.; Buschkuehl, M.; Jonides, J.; and Perrig, W.~J. 2008.
\newblock Improving fluid intelligence with training on working memory.
\newblock \emph{Proceedings of the National Academy of Sciences}, 105(19):
  6829--6833.

\bibitem[{Jahrens and Martinetz(2020)}]{jahrens2020solving}
Jahrens, M.; and Martinetz, T. 2020.
\newblock Solving raven’s progressive matrices with multi-layer relation
  networks.
\newblock In \emph{2020 International Joint Conference on Neural Networks
  (IJCNN)}, 1--6. IEEE.

\bibitem[{LeCun et~al.(1998)LeCun, Bottou, Bengio, and
  Haffner}]{lecun1998gradient}
LeCun, Y.; Bottou, L.; Bengio, Y.; and Haffner, P. 1998.
\newblock Gradient-based learning applied to document recognition.
\newblock \emph{Proceedings of the IEEE}, 86(11): 2278--2324.

\bibitem[{Ma et~al.(2022)Ma, Nie, Yu, Jiang, Xiao, Zhu, Zhu, and
  Anandkumar}]{xiaojianrelvit2022}
Ma, X.; Nie, W.; Yu, Z.; Jiang, H.; Xiao, C.; Zhu, Y.; Zhu, S.; and Anandkumar,
  A. 2022.
\newblock RelViT: Concept-guided Vision Transformer for Visual Relational
  Reasoning.
\newblock In \emph{The Tenth International Conference on Learning
  Representations, {ICLR} 2022, Virtual Event, April 25-29, 2022}.
  OpenReview.net.

\bibitem[{Maguire, Burgess, and O’Keefe(1999)}]{maguire1999human}
Maguire, E.~A.; Burgess, N.; and O’Keefe, J. 1999.
\newblock Human spatial navigation: cognitive maps, sexual dimorphism, and
  neural substrates.
\newblock \emph{Current opinion in neurobiology}, 9(2): 171--177.

\bibitem[{Ma{\l}ki{\'n}ski and
  Ma{\'n}dziuk(2022{\natexlab{a}})}]{malkinski2022deep}
Ma{\l}ki{\'n}ski, M.; and Ma{\'n}dziuk, J. 2022{\natexlab{a}}.
\newblock Deep Learning Methods for Abstract Visual Reasoning: A Survey on
  Raven's Progressive Matrices.
\newblock \emph{arXiv preprint arXiv:2201.12382}.

\bibitem[{Ma{\l}ki{\'n}ski and
  Ma{\'n}dziuk(2022{\natexlab{b}})}]{malkinski2022multi}
Ma{\l}ki{\'n}ski, M.; and Ma{\'n}dziuk, J. 2022{\natexlab{b}}.
\newblock Multi-label contrastive learning for abstract visual reasoning.
\newblock \emph{IEEE Transactions on Neural Networks and Learning Systems}.

\bibitem[{Mao et~al.(2021)Mao, Zhang, Zheng, Ma, Peng, Ding, Zhang, Han
  et~al.}]{mao2021dual}
Mao, M.; Zhang, R.; Zheng, H.; Ma, T.; Peng, Y.; Ding, E.; Zhang, B.; Han, S.;
  et~al. 2021.
\newblock Dual-stream network for visual recognition.
\newblock \emph{Advances in Neural Information Processing Systems}, 34:
  25346--25358.

\bibitem[{Mondal, Webb, and Cohen(2023)}]{mondallearning}
Mondal, S.~S.; Webb, T.~W.; and Cohen, J. 2023.
\newblock Learning to reason over visual objects.
\newblock In \emph{International Conference on Learning Representations}.

\bibitem[{Niebur and Koch(1995)}]{niebur1995control}
Niebur, E.; and Koch, C. 1995.
\newblock Control of selective visual attention: Modeling the" where" pathway.
\newblock \emph{Advances in neural information processing systems}, 8.

\bibitem[{Perfetti et~al.(2009)Perfetti, Saggino, Ferretti, Caulo, Romani, and
  Onofrj}]{perfetti2009differential}
Perfetti, B.; Saggino, A.; Ferretti, A.; Caulo, M.; Romani, G.~L.; and Onofrj,
  M. 2009.
\newblock Differential patterns of cortical activation as a function of fluid
  reasoning complexity.
\newblock \emph{Human brain mapping}, 30(2): 497--510.

\bibitem[{Prabhakaran et~al.(1997)Prabhakaran, Smith, Desmond, Glover, and
  Gabrieli}]{Prabhakaran1997neural}
Prabhakaran, V.; Smith, J.~A.; Desmond, J.~E.; Glover, G.~H.; and Gabrieli,
  J.~D. 1997.
\newblock Neural substrates of fluid reasoning: an fMRI study of neocortical
  activation during performance of the Raven's Progressive Matrices Test.
\newblock \emph{Cognitive psychology}, 33(1): 43--63.

\bibitem[{Raghu et~al.(2021)Raghu, Unterthiner, Kornblith, Zhang, and
  Dosovitskiy}]{raghu2021vision}
Raghu, M.; Unterthiner, T.; Kornblith, S.; Zhang, C.; and Dosovitskiy, A. 2021.
\newblock Do vision transformers see like convolutional neural networks?
\newblock \emph{Advances in Neural Information Processing Systems}, 34:
  12116--12128.

\bibitem[{Rahaman et~al.(2021)Rahaman, Gondal, Joshi, Gehler, Bengio,
  Locatello, and Sch{\"o}lkopf}]{rahaman2021dynamic}
Rahaman, N.; Gondal, M.~W.; Joshi, S.; Gehler, P.; Bengio, Y.; Locatello, F.;
  and Sch{\"o}lkopf, B. 2021.
\newblock Dynamic inference with neural interpreters.
\newblock \emph{Advances in Neural Information Processing Systems}, 34:
  10985--10998.

\bibitem[{Ronneberger, Fischer, and Brox(2015)}]{ronneberger2015u}
Ronneberger, O.; Fischer, P.; and Brox, T. 2015.
\newblock U-net: Convolutional networks for biomedical image segmentation.
\newblock In \emph{Medical Image Computing and Computer-Assisted
  Intervention--MICCAI 2015: 18th International Conference, Munich, Germany,
  October 5-9, 2015, Proceedings, Part III 18}, 234--241. Springer.

\bibitem[{Sahu, Basioti, and Pavlovic(2022)}]{PritishSAViR-T}
Sahu, P.; Basioti, K.; and Pavlovic, V. 2022.
\newblock SAViR-T: Spatially Attentive Visual Reasoning with Transformers.
\newblock \emph{CoRR}, abs/2206.09265.

\bibitem[{Santoro et~al.(2018)Santoro, Hill, Barrett, Morcos, and
  Lillicrap}]{santoro2018measuring}
Santoro, A.; Hill, F.; Barrett, D.; Morcos, A.; and Lillicrap, T. 2018.
\newblock Measuring abstract reasoning in neural networks.
\newblock In \emph{International Conference on Machine Learning}, 4477--4486.

\bibitem[{Santoro et~al.(2017)Santoro, Raposo, Barrett, Malinowski, Pascanu,
  Battaglia, and Lillicrap}]{santoro2017simple}
Santoro, A.; Raposo, D.; Barrett, D.~G.; Malinowski, M.; Pascanu, R.;
  Battaglia, P.; and Lillicrap, T. 2017.
\newblock A simple neural network module for relational reasoning.
\newblock \emph{Advances in neural information processing systems}, 30.

\bibitem[{Simonyan and Zisserman(2014)}]{simonyan2014two}
Simonyan, K.; and Zisserman, A. 2014.
\newblock Two-stream convolutional networks for action recognition in videos.
\newblock \emph{Advances in neural information processing systems}, 27.

\bibitem[{Snow et~al.(1984)Snow, Kyllonen, Marshalek
  et~al.}]{snow1984topography}
Snow, R.~E.; Kyllonen, P.~C.; Marshalek, B.; et~al. 1984.
\newblock The topography of ability and learning correlations.
\newblock \emph{Advances in the psychology of human intelligence}, 2(S 47):
  103.

\bibitem[{Snow and Lohman(1984)}]{snow1984toward}
Snow, R.~E.; and Lohman, D.~F. 1984.
\newblock Toward a theory of cognitive aptitude for learning from instruction.
\newblock \emph{Journal of educational psychology}, 76(3): 347.

\bibitem[{Spratley, Ehinger, and Miller(2020)}]{spratley2020closer}
Spratley, S.; Ehinger, K.; and Miller, T. 2020.
\newblock A closer look at generalisation in raven.
\newblock In \emph{European Conference on Computer Vision}, 601--616. Springer.

\bibitem[{van~der Maaten and Hinton(2008)}]{tsne:vandermaaten08a}
van~der Maaten, L.; and Hinton, G. 2008.
\newblock Visualizing Data using t-SNE.
\newblock \emph{Journal of Machine Learning Research}, 9(86): 2579--2605.

\bibitem[{Vaswani et~al.(2017)Vaswani, Shazeer, Parmar, Uszkoreit, Jones,
  Gomez, Kaiser, and Polosukhin}]{vaswani2017attention}
Vaswani, A.; Shazeer, N.; Parmar, N.; Uszkoreit, J.; Jones, L.; Gomez, A.~N.;
  Kaiser, {\L}.; and Polosukhin, I. 2017.
\newblock Attention is all you need.
\newblock \emph{Advances in neural information processing systems}, 30.

\bibitem[{Wang, Jamnik, and Li{\`{o}}(2020)}]{wang2020abstract}
Wang, D.; Jamnik, M.; and Li{\`{o}}, P. 2020.
\newblock Abstract Diagrammatic Reasoning with Multiplex Graph Networks.
\newblock In \emph{8th International Conference on Learning Representations,
  {ICLR} 2020, Addis Ababa, Ethiopia, April 26-30, 2020}. OpenReview.net.

\bibitem[{Whittington et~al.(2022)Whittington, McCaffary, Bakermans, and
  Behrens}]{whittington2022build}
Whittington, J.~C.; McCaffary, D.; Bakermans, J.~J.; and Behrens, T.~E. 2022.
\newblock How to build a cognitive map.
\newblock \emph{Nature Neuroscience}, 25(10): 1257--1272.

\bibitem[{Whittington et~al.(2020)Whittington, Muller, Mark, Chen, Barry,
  Burgess, and Behrens}]{whittington2020tolman}
Whittington, J.~C.; Muller, T.~H.; Mark, S.; Chen, G.; Barry, C.; Burgess, N.;
  and Behrens, T.~E. 2020.
\newblock The Tolman-Eichenbaum machine: unifying space and relational memory
  through generalization in the hippocampal formation.
\newblock \emph{Cell}, 183(5): 1249--1263.

\bibitem[{Wu et~al.(2020)Wu, Dong, Grosse, and Ba}]{wu2020scattering}
Wu, Y.; Dong, H.; Grosse, R.; and Ba, J. 2020.
\newblock The scattering compositional learner: Discovering objects,
  attributes, relationships in analogical reasoning.
\newblock \emph{arXiv preprint arXiv:2007.04212}.

\bibitem[{Yang et~al.(2023)Yang, You, Zhen, Wang, Wan, Xie, and
  Zhang}]{pmlr-v202-yang23r}
Yang, L.; You, H.; Zhen, Z.; Wang, D.; Wan, X.; Xie, X.; and Zhang, R.-Y. 2023.
\newblock Neural prediction errors enable analogical visual reasoning in human
  standard intelligence tests.
\newblock In \emph{International Conference on Machine Learning}, 39572--39583.
  PMLR.

\bibitem[{Zhang et~al.(2019{\natexlab{a}})Zhang, Gao, Jia, Zhu, and
  Zhu}]{zhang2019raven}
Zhang, C.; Gao, F.; Jia, B.; Zhu, Y.; and Zhu, S.-C. 2019{\natexlab{a}}.
\newblock Raven: A dataset for relational and analogical visual reasoning.
\newblock In \emph{Proceedings of the IEEE/CVF Conference on Computer Vision
  and Pattern Recognition}, 5317--5327.

\bibitem[{Zhang et~al.(2019{\natexlab{b}})Zhang, Jia, Gao, Zhu, Lu, and
  Zhu}]{zhang2019learning}
Zhang, C.; Jia, B.; Gao, F.; Zhu, Y.; Lu, H.; and Zhu, S.-C.
  2019{\natexlab{b}}.
\newblock Learning perceptual inference by contrasting.
\newblock \emph{Advances in Neural Information Processing Systems}, 32.

\bibitem[{Zhang et~al.(2021)Zhang, Jia, Zhu, and Zhu}]{zhang2021abstract}
Zhang, C.; Jia, B.; Zhu, S.-C.; and Zhu, Y. 2021.
\newblock Abstract spatial-temporal reasoning via probabilistic abduction and
  execution.
\newblock In \emph{Proceedings of the IEEE/CVF Conference on Computer Vision
  and Pattern Recognition}, 9736--9746.

\bibitem[{Zhang et~al.(2022{\natexlab{a}})Zhang, Xie, Jia, Wu, Zhu, and
  Zhu}]{zhang2021learning}
Zhang, C.; Xie, S.; Jia, B.; Wu, Y.~N.; Zhu, S.-C.; and Zhu, Y.
  2022{\natexlab{a}}.
\newblock Learning algebraic representation for systematic generalization in
  abstract reasoning.
\newblock In \emph{European Conference on Computer Vision}, 692--709. Springer.

\bibitem[{Zhang et~al.(2022{\natexlab{b}})Zhang, Tang, Mo, Liu, and
  Song}]{zhang2022learning}
Zhang, W.; Tang, L.; Mo, S.; Liu, X.; and Song, S. 2022{\natexlab{b}}.
\newblock Learning Robust Rule Representations for Abstract Reasoning via
  Internal Inferences.
\newblock In \emph{Advances in Neural Information Processing Systems}.

\bibitem[{Zheng, Zha, and Wei(2019)}]{zheng2019abstract}
Zheng, K.; Zha, Z.-J.; and Wei, W. 2019.
\newblock Abstract reasoning with distracting features.
\newblock \emph{Advances in Neural Information Processing Systems}, 32.

\bibitem[{Zhou et~al.(2021)Zhou, Zhou, Zhou, and Li}]{zhou2021spatial}
Zhou, H.; Zhou, W.; Zhou, Y.; and Li, H. 2021.
\newblock Spatial-temporal multi-cue network for sign language recognition and
  translation.
\newblock \emph{IEEE Transactions on Multimedia}, 24: 768--779.

\bibitem[{Zhuo and Kankanhalli(2020)}]{zhuo2020effective}
Zhuo, T.; and Kankanhalli, M. 2020.
\newblock Effective abstract reasoning with dual-contrast network.
\newblock In \emph{International Conference on Learning Representations}.

\bibitem[{Zolfaghari et~al.(2017)Zolfaghari, Oliveira, Sedaghat, and
  Brox}]{zolfaghari2017chained}
Zolfaghari, M.; Oliveira, G.~L.; Sedaghat, N.; and Brox, T. 2017.
\newblock Chained multi-stream networks exploiting pose, motion, and appearance
  for action classification and detection.
\newblock In \emph{Proceedings of the IEEE International Conference on Computer
  Vision}, 2904--2913.

\end{thebibliography}

\end{document}